\documentclass[runningheads]{llncs}
\usepackage{amsmath}
\usepackage{algorithm}
\usepackage{algorithmic}
\usepackage{hyperref}
\usepackage{graphicx}
\begin{document}
\title{ZJUNlict Extended Team Description Paper for RoboCup 2019}
%
%
\author{
Zheyuan Huang\inst{1} \and
Lingyun Chen\inst{1} \and
Jiacheng Li\inst{1} \and
Yunkai Wang\inst{1} \and
Zexi Chen\inst{1} \and
Licheng Wen\inst{1} \and
Jianyang Gu\inst{1} \and
Peng Hu\inst{1} \and
Rong Xiong\inst{1}
}
\authorrunning{ }
%
\institute{Zhejiang University, Zheda Road No.38, Hangzhou, Zhejiang Province, P.R.China
\email{rxiong@iipc.zju.edu.cn}\\
\url{https://zjunlict.cn}}
\maketitle              
\begin{abstract}
For the Small Size League of RoboCup 2018, Team ZJUNLict has won the champion and therefore, this paper thoroughly described the devotion which ZJUNLict has devoted and the effort that ZJUNLict has contributed. There are three mean optimizations for the mechanical part which accounted for most of our incredible goals, they are ``Touching Point Optimization'', ``Damping System Optimization'', and ``Dribbler Optimization''. For the electrical part, we realized ``Direct Torque Control'', ``Efficient Radio Communication Protocol'' which will be credited for stabilizing the dribbler and a more secure communication between robots and the computer. Our software group contributed as much as our hardware group with the effort of ``Vision Lost Compensation'' to predict the movement by kalman filter, and ``Interception Prediction Algorithm'' to achieve some skills and improve our ball possession rate.

\end{abstract}
\section{ZJUNlict New Dribbler Design}
\subsection{Typical Dribblers and Existing Problems}
\begin{figure}[h!t]
\centering
\begin{minipage}{.5\textwidth}
  \centering
  \includegraphics[height=1.5in]{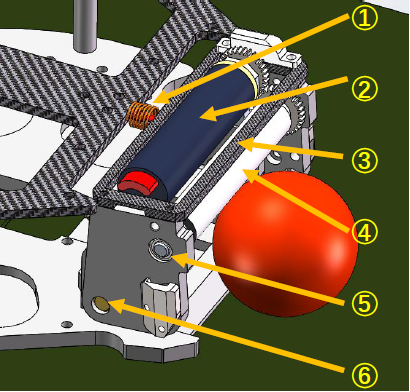}
  \caption{Typical Dribbler (1.Unidirectional Damper 2.Dribble Motor 3.Connect Shelf 4.Dribbling-bar 5.Side Plate 6.Rotational Joint)}
  \label{dribbler}
\end{minipage}%
\begin{minipage}{.5\textwidth}
  \centering
  \includegraphics[height=1.5in]{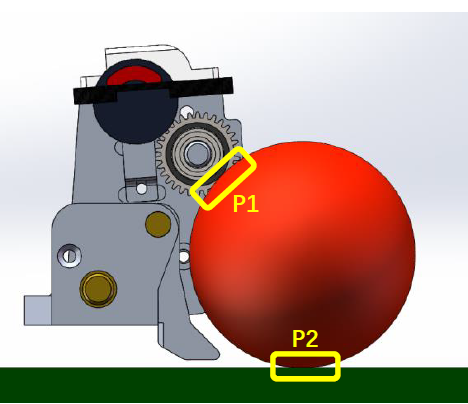}
  \caption{2-touch-point Model}
  \label{2-touch-point}
\end{minipage}
\end{figure}

The small size league robots do not really have foot like human beings. Instead, they have dribblers. A dribbler is a device that can help dribble and catch the ball. As shown in Fig. \ref{dribbler}, a typical dribbler has the following features. A shelf connects 2 side plates and the dribbling motor is fixed on one side plate. Between the 2 side plates is a cylindrical dribbling-bar driven by the dribbling motor. The whole device has only one degree of freedom of rotation and the joints are fixed on the robot flame. Usually there is a unidirectional spring-damping system locates between the shelf and the robot frame to help improve the stability of dribbling as well as absorbing the energy when catching the ball. The dribbling-bar driven by the dribbling motor provides torque to make the ball spin backward when the contact between the ball and dribbling-bar exits so that the ball can be ‘locked’ by this device in ideal conditions. And the carpet provides supporting force and frictional force and therefore there are 2 touch points on the ball and in this paper we called it a 2-touch-point model (Fig. \ref{2-touch-point}). For the motor control, most teams try to keep the dribbling-bar at a constant rotational speed when dribbling the ball and therefore it is actually an open loop control mode for dribbling. Unfortunately, this 2-touch-point dribbler with unidirectional spring-damping system and passive control mode does not provide ideal dribbling performances. It is quite easy for the ball to bounce back and forth when launching the dribbling motor. The device might also not absorb enough kinetic energy of the moving ball when catching it so it will bounce back and there occurs a catching failure. Actually it is also hard to greatly improve its performance by simply changing the material of dribbling-bar, adjusting the damping and stiffness of the spring-damping system or adjusting the rotational speed of motor. This structure has natural defects with passive control mode.

\begin{figure}[h!t]
  \centering
  \includegraphics[width=3in]{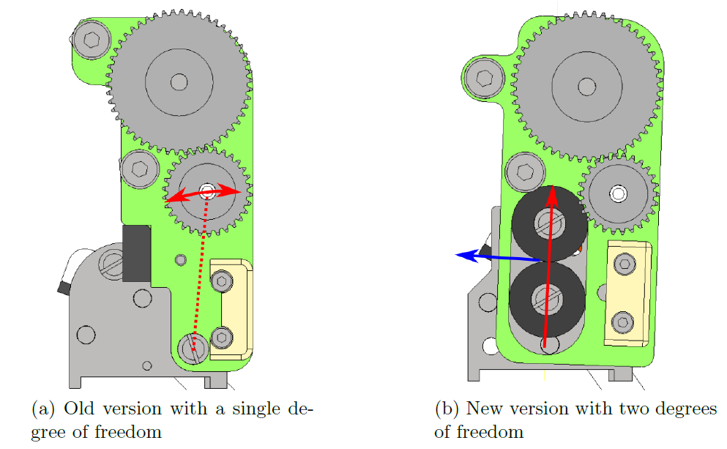}
  \caption{Energy Absorbing by Creating a Dual Freedom \cite{tiger_2018}}
  \label{Tigers}
\end{figure}

Tigers\cite{tiger_2018} developed a dribbler with 2 degree of freedom (Fig. \ref{Tigers}). Except for rotational degrees of freedom, the side plates can slide up and down along two damped linear guides with screws covered by thick silicon ring, by doing this, much more kinetic energy will be absorbed by the silicone ring as well as be transferred into the potential energy of the device when catching the ball. It was approved that this device worked quite well with catching and dribbling in static conditions. For example, when the robot stays stilly or just moves back and forth slowly, catching a ball with coming speed up to 5m/s is quite easy. But considering the real competition environment, the condition will not be that idealistic and more complex movements are needed, indeed. For example, when two robots scramble for a ball, we want our robot able to turn around while dribbling so that it can make space for passing. Also when all shot space is blocked by defenders we want our robot able to do some actions like moving laterally while dribbling to create space to score. In a word, a stronger dribbler is urgently in need.

\subsection{Mechanical Improvements}
\begin{figure}[h!t]
\centering
\begin{minipage}{.5\textwidth}
  \centering
  \includegraphics[height=1.5in]{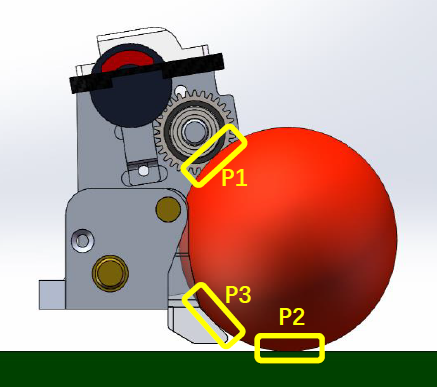}
  \caption{3-touch-point Model}
  \label{3-touch-point}
\end{minipage}%
\begin{minipage}{.5\textwidth}
  \centering
  \includegraphics[height=1.5in]{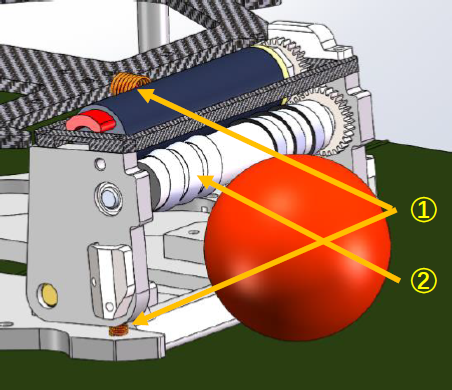}
  \caption{New Damper (1.Bidirectional Damper 2.Screw Dribbling Bar)}
  \label{new-damper}
\end{minipage}
\end{figure}

Considering the purpose above, we devoted ourselves on the dribbler. Firstly, we adjust the geometry parameters of the device so that the ball can touch the chip shovel in steady state, which means both carpet and chip shovel can provide supporting force and frictional force to the ball so we called it a 3-touch-point model (Fig. \ref{3-touch-point}). Hopefully this design will limit the bouncing space and it will be much easier for the coming ball to enter a steady state. In addition, we found that there will be a hard contact between the side plates and the baseplate when the dribbler hits the baseplate. So besides the foam between the shelf and the robot frame, we stick 1.5 mm thick tape between the side plates and baseplate so there will be a soft contact when the dribbler hits the baseplate. Actually this design makes up a bidirectional spring-damping system (Fig. \ref{new-damper}) and improves the dynamic behavior of the dribbler. Hopefully it can reduce the bouncing amplitude of the ball when dribbling as well as absorbing more kinetic energy when catching the ball. To improve the dribbling performance when the robot rotates or moves laterally, we also made a dribbling-bar with screw using 3D printing rubber so that it can provide lateral force to the ball when dribbling as shown in Fig.\ref{new-damper}. With the structural innovation above, we create a quite good passive control dribbler. Both simulation and experimental work were done to verify the mechanical improvement.

\subsection{Tests and Verifications}
According to the catching ability tests, the typical 2-touch-point dribbler with unidirectional spring-damping system could catch a ball with coming speed up to $3 m/s$ and the new 3-touch-point dribbler with bidirectional spring-damping system could catch a ball with coming   speed up to $8.5 m/s$. The results were quite clear that the new dribbler has better dribbling and catching ability.
In addition, we made simple tests to see the effect of screw added on the dribbling bar. The dribbling motor was launched and after the dribbling entering the steady state, we made the robot spin around. The rotational acceleration is $20 d/s^2$ and the rotation speed was recorded at the time the ball left the dribbler. This simple test was carried out 10 times for both smooth dribbling-bar and screw dribbling-bar, which were made by some same material. As show in Tabel.1 below, the average escape speed of smooth dribbling-bar is $402 d/s$ and for the screw dribbling-bar is $622 d/s$. So it was proved that the design of screw could improve the dynamic dribbling performance of dribbler.

\begin{table}
\centering
\caption{Dynamic Dribbling Ability Comparison Between Smooth Dribbling-bar and Screw Dribbling-bar}
\label{tab1}
\begin{tabular}{| c | c | c | c | c | c | c | c | c | c | c | c |}
\hline
Dribbling-bar Type & 1 & 2 & 3 & 4 & 5 & 6 & 7 & 8 & 9 & 10 & Average\\
\hline
Smooth Dribbling-bar $(d/s)$ & 400 & 340 & 380 & 360 & 380 & 420 & 400 & 420 & 400 & 520 & 402\\
\hline
Screw Dribbling-bar $(d/s)$ & 600 & 580 & 580 & 580 & 620 & 680 & 620 & 680 & 640 & 640 & 622\\
\hline
\end{tabular}
\end{table}

\subsection{Mechanical Simulation and Mechanism Exploration}
\begin{figure}[h!t]
  \centering
  \includegraphics[width=3.5in]{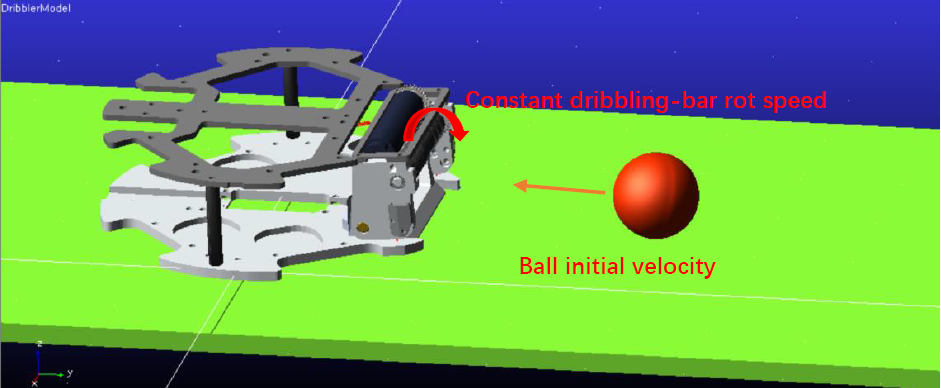}
  \caption{Simulation Environment in ADAMS}
  \label{ADAMS}
\end{figure}

\begin{figure}[h!t]
  \centering
  \includegraphics[width=4in]{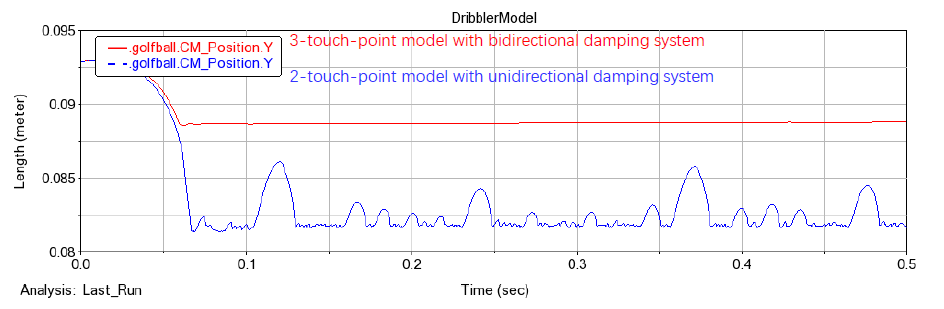}
  \caption{Results of Ball Position Comparison}
  \label{ball-position}
\end{figure}

\begin{figure}[h!t]
  \centering
  \includegraphics[width=3in]{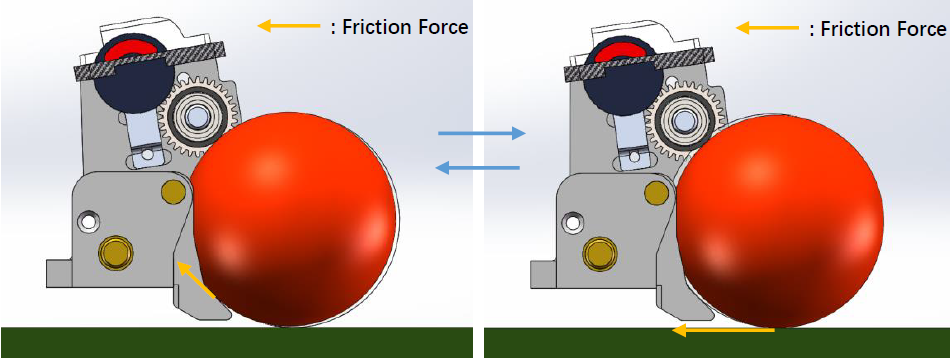}
  \caption{Dynamic Steady State of 3-touch-point Model}
  \label{D-3-touch-point}
\end{figure}

\begin{figure}[h!t]
  \centering
  \includegraphics[width=3in]{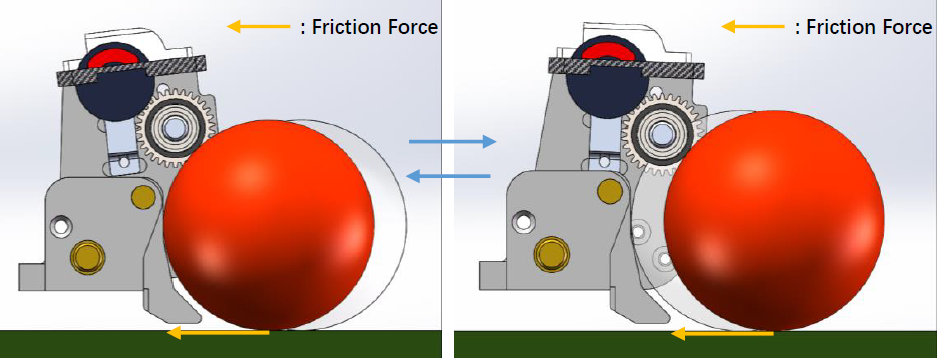}
  \caption{Dynamic Steady State of 2-touch-point Model}
  \label{D-2-touch-point}
\end{figure}

In order to explore the mechanism of the improvements, simulation models of 3-touch-point model with bidirectional spring-damping system compared with the 2-touch-point model with unidirectional spring-damping system were built in ADAMS.
A constant rotation speed of dribbling-bar with $3300r/min$ was given and the ball was released with initial speed of $0.1m/s$ to hit the dribbler (Fig. \ref{ADAMS}). From the simulation results of ball positions (Fig. \ref{ball-position}), the dribbling of the 3-touch-point model was significantly more stable than that of the 2-touch-point-model. It could also be seen that there is no strict steady state, the ball will keep bouncing back and forth and we judge the steady state by the bouncing amplitude which means, if the bouncing amplitude is small enough that the ball never bounces off the dribbler, we can judge it as a steady state. The result also explained why a 3-touch-point model is better than a 2-touch-point model. Normally the dynamic friction coefficient between the ball-carpet surface is greater than that of the ball-chip shove surface. Therefore, when the ball driven by the dribbling-bar moves from the carpet on to the chip shove surface, there will be a sudden drop of frictional force, and the ball will be pushed back on the carpet. And once the ball touches the carpet, there will be a sudden increase of frictional force, the ball will be driven onto the chip shove again. In this kind of state, the amount of spring compression will not change much so that the dribbling system will enter a periodical dynamic steady state (Fig. \ref{D-3-touch-point}). In contrast, with a 2-touch-point system, the friction force will not change much so the ball will enter much more into the dribbler and there will be a bigger compression of the spring-damping system. Therefore the ball will also be pushed back more and totally the bouncing amplitude will be much greater, or even the ball will bounce off the dribbler(Fig. \ref{D-2-touch-point}).
The key to make this device better is to change the passive control mode to active control mode. Adaptive torque controlling model by adjusting the torque of dribbling motor in real time was developed.

\section{Electrical System}

\subsection{Overview}

For a typical Small Size League robot, the electrical system is responsible for the overall control of the hardware based on commands sent by a team's strategy program. The detailed tasks for the electrical system include driving motors, wireless communication, charging and discharging capacitors for shooting and chipping. There are three configurations of electrical systems shown in the Table \ref{Elec_P1} below currently used in ZJUNlict robots. The single Cyclone III FPGA configuration was adopted since 2012 and explained in 2013 and 2014 champion papers \cite{zju_champion_2013,zju_champion_2014}. The FPGA handles both motor control and other tasks such as communication and motion sensor fusion based on embedded Nios II processor. The micro-controller STM32F407 was added to take over tasks other than motor control since late 2017 \cite{zju_2018}. Since 2018, a single micro-controller STM32H743 capable of operation frequency up to 400MHz combined with five BLDC controller Allegro A3930 was able to handle all the tasks. Other improvements include increasing each encoder's counts per revolution (CPR) to increase motor low-speed control performance, implementation of the accelerometer and compass to achieve more accurate motion tracking and switching to nRF24L01+ wireless IC to deliver higher bandwidth communication with better signal sensitivity. The PCB designs and related firmware shown in the Table \ref{Elec_P1} have been open source on \href{https://github.com/ZJUNlict/Electronics}{Github}.

\begin{table}[h!t]
    \centering
    \caption{ZJUNlict Electrical System Configurations}
    \label{Elec_P1}
    \includegraphics[height=2in]{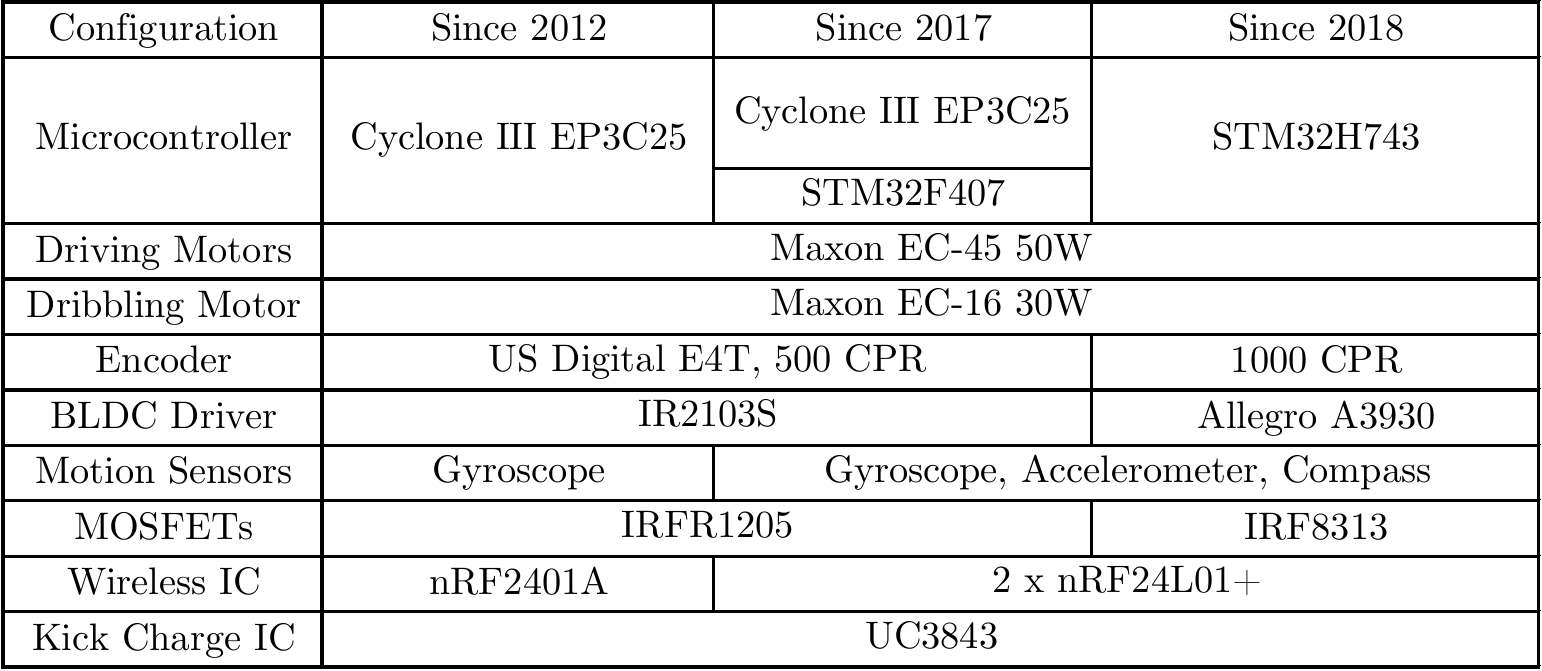}
\end{table}

\subsection{Efficient Radio Communication Protocol}

The communication quality becomes a pressing problem when the competition requires a larger field and more robots. For a typical nRF24L01+ wireless package (maximum 32 bytes) consists of 5 bytes receiver address, 2 bytes CRC and 25 bytes user payload. Compared with other teams' communication protocol \cite{tiger_2016,tiger_2018} which often requires one package for each robot, the presented efficient communication protocol is shown in the Table \ref{Elec_P2} significantly reduces the package amount required in each control period. Apart from 1 byte package header indicates the package type and robot existence in each package, each robot takes up to 6 bytes for the velocity, dribbler and kick command. For 25 bytes user payload, each package is able to control up to 4 robots. For a typical 60Hz control frequency based on SSL-vision information, the minimal data rate can be calculated as 30.72kbits/s. So the bandwidth of nRF24L01+ can be set to as low as 1Mbits/s or even 250kbits/s which requires fewer frequency bandwidth and increase the signal sensitivity based on nRF24L01+ chip datasheet \cite{nrf24L01}. By reducing the communication traffic in each control period reduces the possibility of radio interference.
\vspace{-0.4cm}
\begin{table}[h!t]
    \centering
    \caption{ZJUNlict Radio Communication Protocol}
    \label{Elec_P2}
    \includegraphics[height=1.7in]{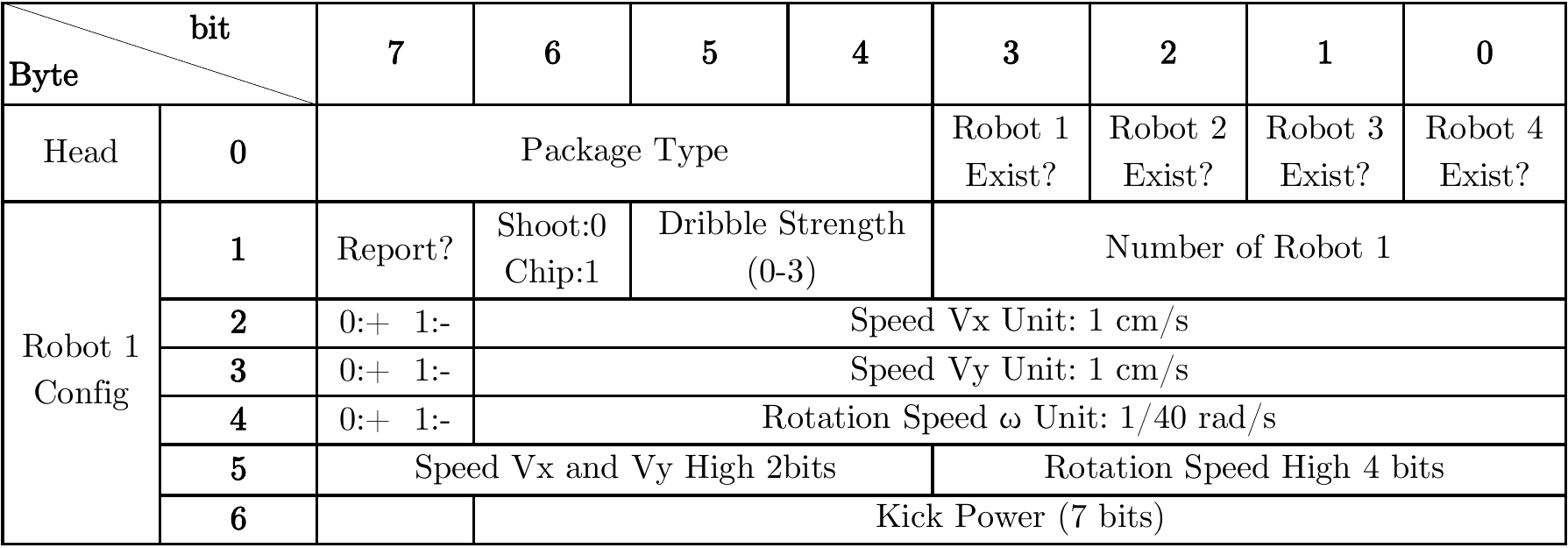}
\end{table}

\section{SSL Vision Solution}
\subsection{Existing Problems}
The existing image recognition system of SSL is shot by cameras ($780\times580$ YUV422 $60Hz$) which suspended about 4m above the field. After image acquisition, the vision software provided by SSL official performs color block recognition progress on the ball (orange) and the color code on the top of each robot. The software determines the robot's information (team, id, orientation) based on the color code combination at the top of robot, and recognizes the position of the ball based on the orange color patch. Finally the robot and ball information is transmitted to our program for processing in the form of UDP packets.

\begin{figure}[h!t]
  \centering
  \includegraphics[width=3in]{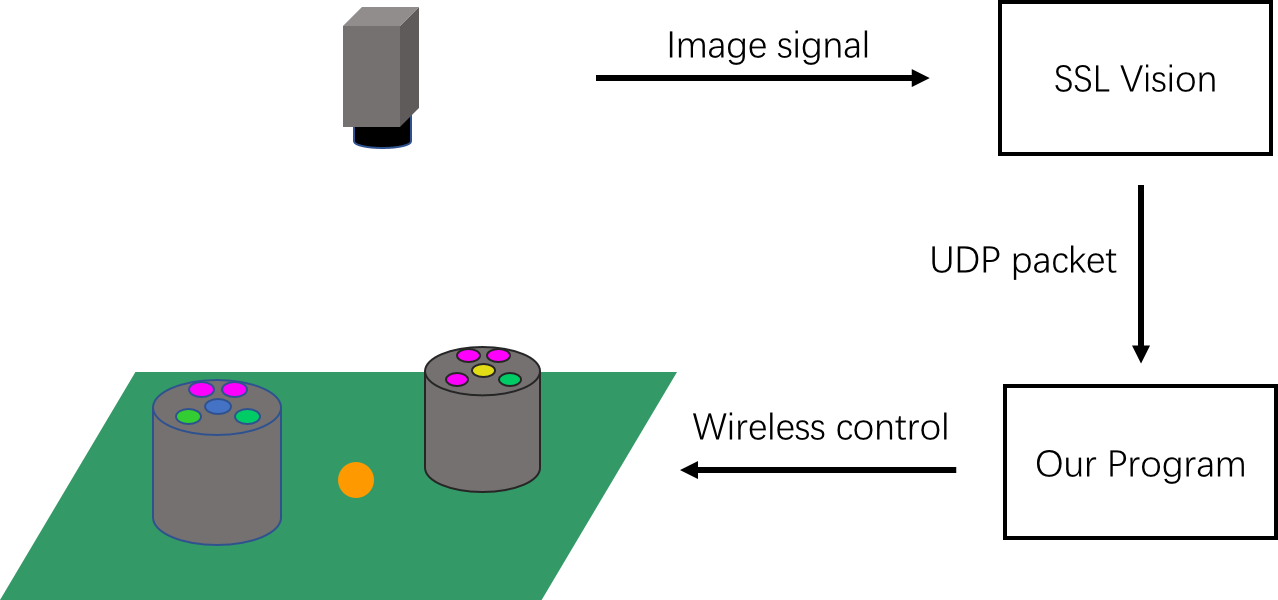}
  \caption{Vision System Introduction}
  \label{vision-intro}
\end{figure}

As the picture below(Fig. \ref{vision-intro}) shows the basic process of the whole SSL vision system.
This image system has been used in the SSL competition for around ten years. As the size of the field continues to enlarge, the number of cameras on the field has also increased from 2 to 8(in this year's competition).  Using color block recognition algorithm accordingly will cause the image processed by the graphic processor to survive the following problems:

\subsubsection{Noise in the position information}
Taking ball as an example, the official vision software recognizes a rectangular orange area as a ball. Therefore, even if the ball itself does not move, the rectangular orange area determined by each frame might still be different, resulting in a small range of jitter in the position of the software recognition.Similarly, there is also jitter in the position of the robot.

\begin{figure}[h!t]
  \centering
  \includegraphics[width=3in]{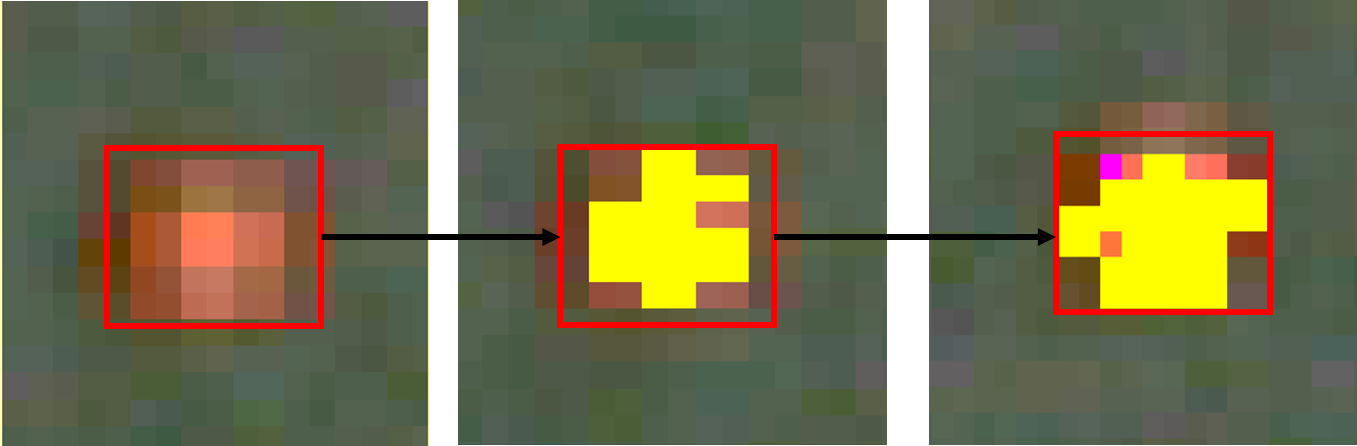}
  \caption{Ball Recognition in Two Different Frame}
  \label{ball-unstable}
\end{figure}

Take Fig. \ref{ball-unstable} as an example. Fig. \ref{ball-unstable}.1 is the original image captured by camera. However, as Fig. \ref{ball-unstable}.2 and Fig. \ref{ball-unstable}.3 shows, the recognized color block varies from frame to frame.

\subsubsection{Light Interference}
The threshold range of various colors in the vision software needs to manually set, and the difference in light environment will directly affect the performance of different colors recognition. As a result, the official software only works properly in a relatively specific light environment (generally a field lighting with stable brightness). Once the light changes beyond the limit, it needs to manually set the color threshold again.

\subsubsection{Object Missing}
Although the camera is overlooking the scene, there will still be cases where the object is lost in the vision. For example, when the ball is moving at high speed, the color of the ball captured by the camera will become lighter and will form a ``fading'' phenomenon, which causes the camera fails to recognize the ball. Fig. \ref{high-speed-ball} clearly shows the vision output when the ball is moving at a high speed.

In some cases, when the robot takes the ball or while two robots are competing for the ball, the camera will not be able to capture the ball because the robot's body will block the ball, which account for image loss.

\subsubsection{Image Recognition Error}
\begin{figure}[h!t]
\centering
\begin{minipage}{.5\textwidth}
  \centering
  \includegraphics[height=1.0in]{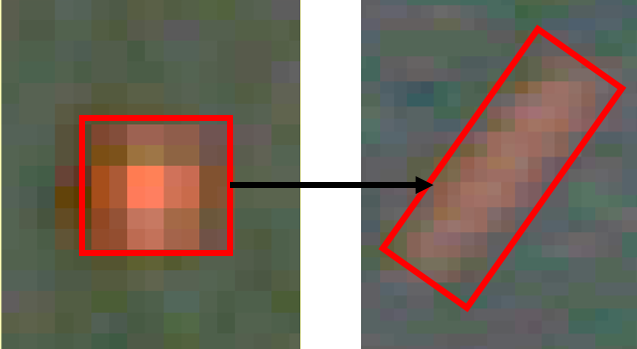}
  \caption{Ball Recogintion in High Velocity}
  \label{high-speed-ball}
\end{minipage}%
\begin{minipage}{.5\textwidth}
  \centering
  \includegraphics[height=1.0in]{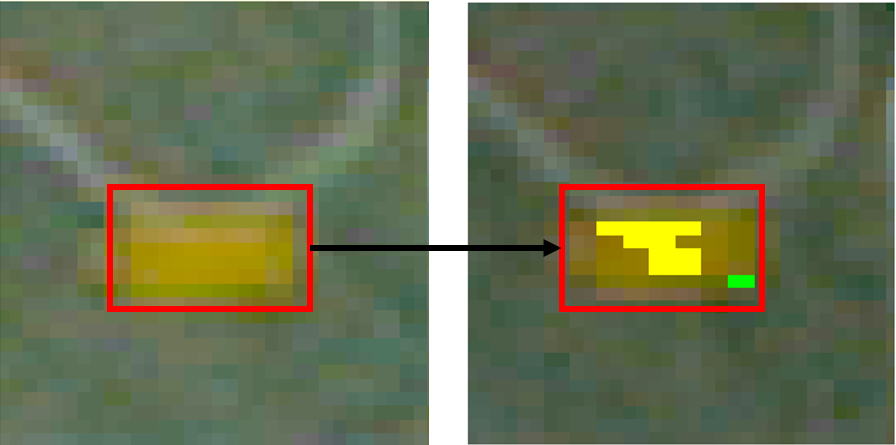}
  \caption{A Non-ball Object Recognize as a Ball}
  \label{invalid-ball-detect}
\end{minipage}
\end{figure}

In some cases within a the game, the person's skin color is similar to orange. Therefore the software will recognize a human skin as a ball, thus increasing the wrong information(Fig. \ref{invalid-ball-detect}). When the robot is located at the edge of the camera's coverage, there is a severe image distortion, and the recognition accuracy of the color code is further reduced, and problems such as unrecognizable robot or robot direction recognition errors might occur.

\subsubsection{Tracking from Multiple Overlapping Cameras}
For up to eight cameras, multiple cameras are visible in many areas of the site (maximum of 4). Due to differences in camera parameters and distortion, the position of the same object in different cameras is different.

In summary, since the official image recognition software of SSL does not provide us with images of sufficient accuracy we need, we need to process these location information. To this end, we have come up with a complete set of vision solutions.

\subsection{Solution Introduction}
Our image solution provides a code framework that covers the various special cases described above, allowing us to perform algorithmic processing for each situation. After receiving the UDP packet sent by the Graphic Processor, the program will automatically judge the current image quality and suspicious conditions for subsequent algorithm processing.

\subsubsection{Noise Cancellation}
For raw data containing noise, we use a Kalman filter considering noise cancellation.

In order to use the Kalman filter to estimate the internal state of a process given only a sequence of noisy observations, one must model the process in accordance with the framework of the Kalman filter. This means specifying the following matrices:

\begin{itemize}
\item
  \(F_k\), the state-transition model;
\item
  \(H_k\), the observation model;
\item
  \(Q_k\), the covariance of the process noise;
\item
  \(R_k\), the covariance of the observation noise;
\item
  and sometimes \(B_k\), the control-input model, for each time-step,
  \emph{k}, as described below.
\end{itemize}

The Kalman filter model assumes the true state at time $k$ is evolved from the state at $(k-1)$ according to
\begin{equation}
{{x} _{k}={F} _{k}{x} _{k-1}+{B} _{k}{u} _{k}+{w} _{k}}
\end{equation}

At the same time, because the data noise is effectively eliminated after Kalman filtering, we can also rely on these data for velocity estimation and position prediction.

\subsubsection{Object Confidence}
In order to solve the misjudgment and missed information of the original image itself, we maintaind the confidence of the ball and the robot on the field.
\begin{equation}
P_{o,t}=P_{o,t-1}+P(seen,t)-P(lost,t), 0\leq P_o\leq 1
\end{equation}
The above is the mathematical expression of confidence. among them:
\begin{itemize}
\item
  \(P_{o,t}\) is the confidence of the object o at time t.
\item
  \(P(seen,t)\) is the probability rise constant of the object o
  appearing in the image at time t
\item
  \(P(lost,t)\) is the probability reduction coefficient of the object o
  disappearing on the image at time t
\end{itemize}
According to the above formula, we set a confidence threshold of $P_v$, then
\begin{equation}
object=\left\{
\begin{aligned}
& valid, &{P_{o,t}>P_v} \\
& invalid, &{P_{o,t} \leq P_v}
\end{aligned}\right.
\end{equation}
This solution effectively eliminates the effects of loss of objects due to cameras, light, and the like. At the same time, interference caused by similar objects such as skin is not considered a valid object because its duration is short and its confidence is lower than the confidence threshold.

\subsubsection{Camera Parameter Identification}
Due to the complexity of multiple camera coverage areas on the site, we have adopted an algorithm that automatically identifies camera parameters. While continuously receiving image information, we continuously calculate and update the coverage area, parameters, etc. of the camera.

When an object appears in the field of view of multiple cameras at the same time, we will calculate its actual position by the following formula: $\bar r_{real} $

\begin{equation}
{\bar r_{real}}={\displaystyle\frac  {\sum _{{i=1}}^{k}\displaystyle\frac {\bar {r_i}-{\bar r_{cam_i}}}{R_{cam_i}}r_{i}}{\sum _{{i=1}}^{k}\displaystyle\frac {\bar {r_i}-{\bar r_{cam_i}}}{R_{cam_i}}}}
\end{equation}

Among them
\begin{itemize}
\item
  \(\bar r_{cam_i}\) is the projection coordinate of camera i
\item
  \(R_{cam_i}\) is the coverage distance of camera i
\item
  k is the number of cameras that can see the current object
\end{itemize}

\subsection{Results}
We use the simulation software grSim to test the actual effect of our image module.  Grsim can adjust the noise (Gaussian noise) and packet loss rate of the original output image to simulate the effect of real games.

We use the pass success rate to reflect the accuracy of our image module handlers, and we will test the success rate of 100 passes in the current environment.


%
%

\begin{figure}[h!t]
\centering
\begin{minipage}{.5\textwidth}
  \centering
  \includegraphics[height=1.5in]{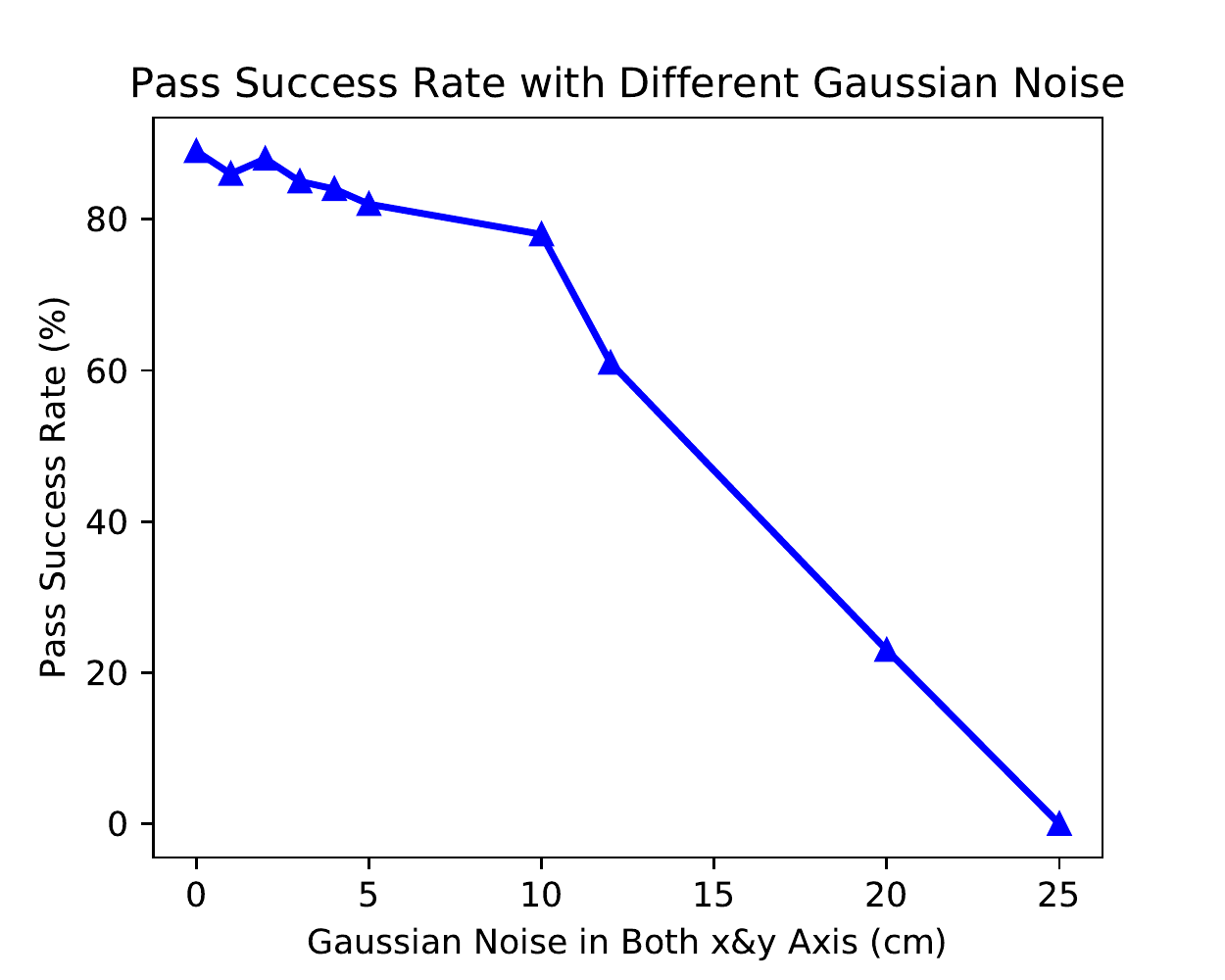}
  \caption{Pass Success Rate with Different Gaussian Noise in Both x and y Axis}
  \label{GN}
\end{minipage}%
\begin{minipage}{.5\textwidth}
  \centering
  \includegraphics[height=1.5in]{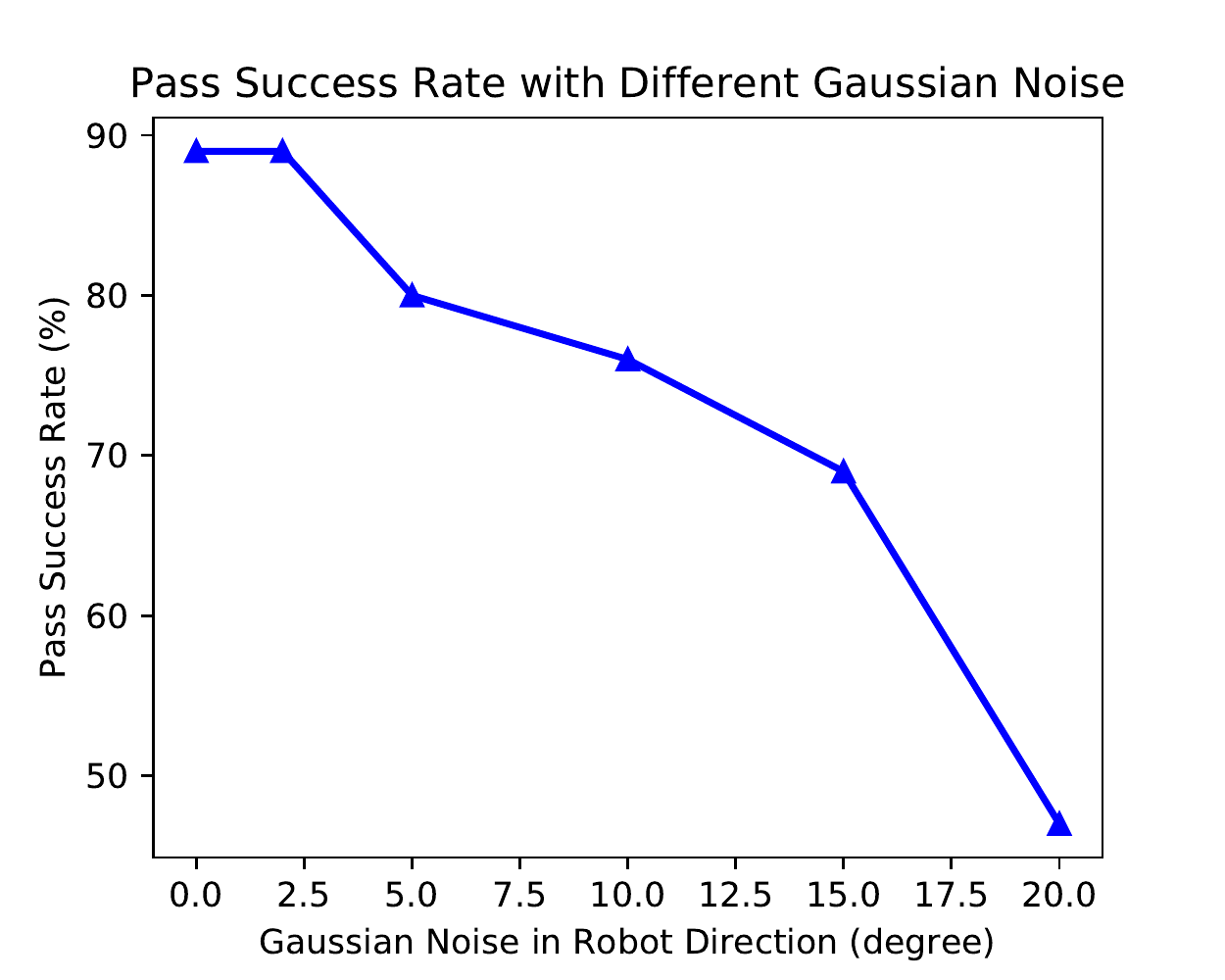}
  \caption{Pass Success Rate with Different Gaussian Noise in Robot Direction}
  \label{GN-Angle}
\end{minipage}
\end{figure}

\begin{figure}[h!t]
  \centering
  \includegraphics[height=1.5in]{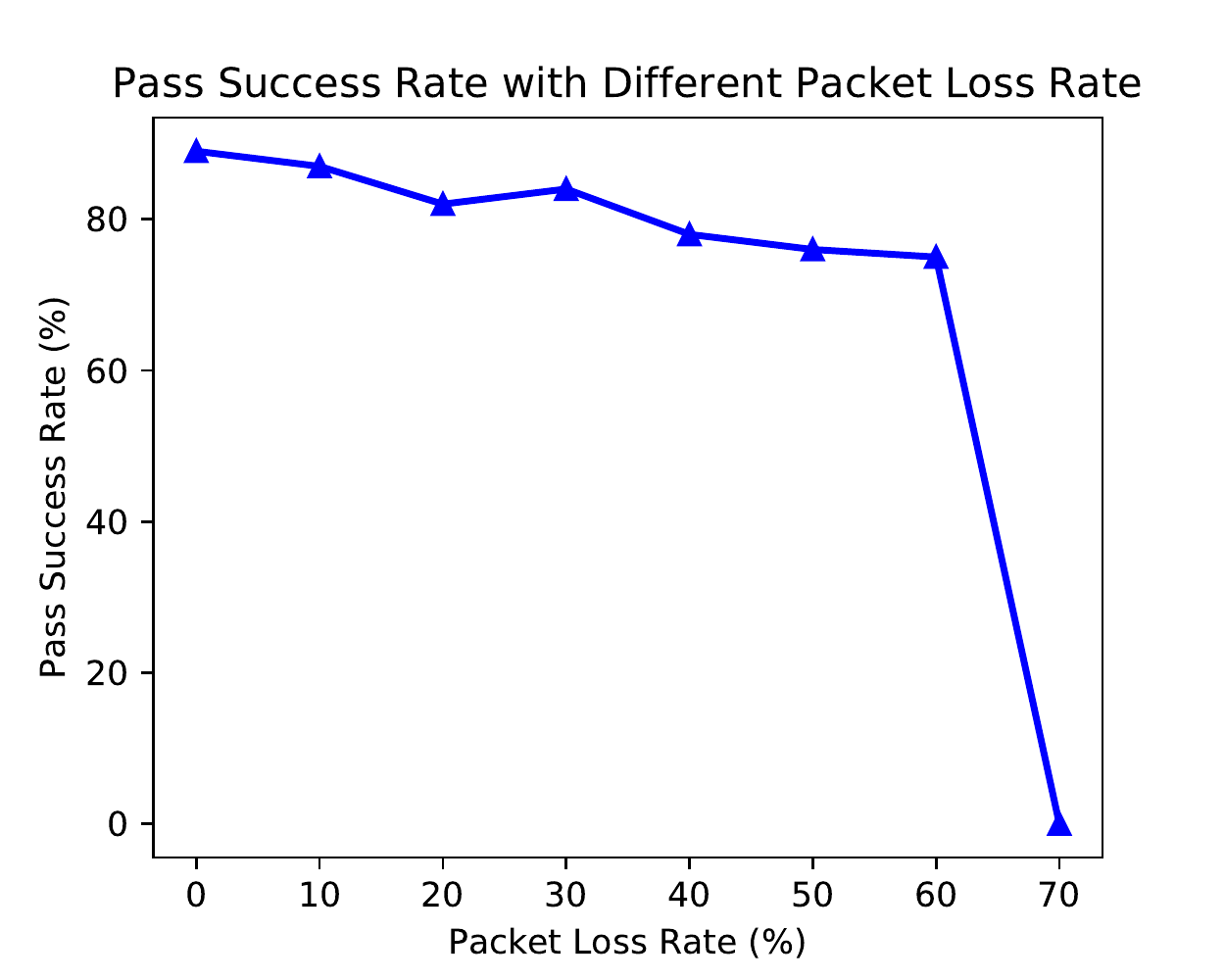}
  \caption{Pass Success Rate with Different Packet Loss Rate}
  \label{Loss}
\end{figure}

\section{Interception Prediction Algorithm and Application}
\subsection{Robot Arrival Time Prediction}
In our system, we adopted the method used in \cite{cmu}. First, we carried out RRT global planner, and then the velocity planner based on the path points generated by RRT. For the velocity planner, we use trapezoidal programming. Since it is the omnidirectional wheel that we used, we decompose the translational speed and rotational speed into a 2d planner and a 1d planner. Then, we decompose the translational velocity into two directions with the orientation from the starting location to the target location as the X-axis, which is beneficial for the robot to achieve the maximum velocity in the x-direction, while the velocity in the y-direction decreases to zero as soon as possible. This will reduce the coupling between the two directions. Therefore, we're basically doing three 1d planner, and then combine them together. For each 1d planner, we will use the maximum acceleration and maximum deceleration under ideal conditions to make a trapezoidal program. Therefore, it could reach the target location with the optimal time, which we can accurately predict.

\subsection{Search-Based Interception Prediction Algorithm}
On the basis of realizing the algorithm of accurately predicting the robot's arrival time to a certain destination, we developed a search-based algorithm that predicts the shortest interception time and the best interception point for the robots. In one actual game, according to the movement ability of both sides, we will make an interception prediction for each robot on both sides of the field in each frame. This is very important for the realization of our single robot skills and the realization of multi-robot attack and defense conversion.
In order to ensure the feasibility and real-time of this process, we use the search-based strategy to search the time at equal intervals with a fixed minimum interval of $\Delta t$(such as 1/60 of a second). At a certain moment $t$, we obtain the location and speed of both the robot and the ball, then calculate the location$P_i$ that the ball can reach at any time $t+i\Delta t (i=0,1,2,3,...)$ in the future under the action of the frictional force of the field in a straight line motion with uniform deceleration(the acceleration of the ball can be obtained according to the measured friction coefficient of the field). Then, starting from $i=0$, it traversed the search points to predict the time $T_i$ that would take a robot to reach the point$P_i$. If $T_k \leq t+k\Delta t$ is satisfied after the k-th interval, point $P_k$ is considered to be the best interception point $P_{best}$ of the robot, and $T_k$ is the shortest interception time $T_{best}$ of the robot. Algorithm. \ref{alg} shows the specific algorithm pseudocode.

There are two extreme cases. One is that the ball has already stopped before the robot intercepts the ball. At this time, the location where the ball stops is the optimal interception location, and the time when the robot reaches the location is the shortest interception time. Another is that the ball has been out of bounds before the robot intercepted it. At this time, in order to ensure that the algorithm can always get a solution, we take the out of bounds location as the best interception location, and the time to the out of bounds location as the shortest interception time. If the prediction of interception time is relatively conservative, such as adding fixed adjustment time $T_m$ to the predicted robot arrival time $T_i$, to ensure a higher success rate of the robot to intercept the ball. It will be found in the actual application that the robot will run more directly to the boundary to intercept the ball.

\begin{figure}[h!t]
\centering
\begin{minipage}{.5\textwidth}
  \centering
  \includegraphics[height=1.5in]{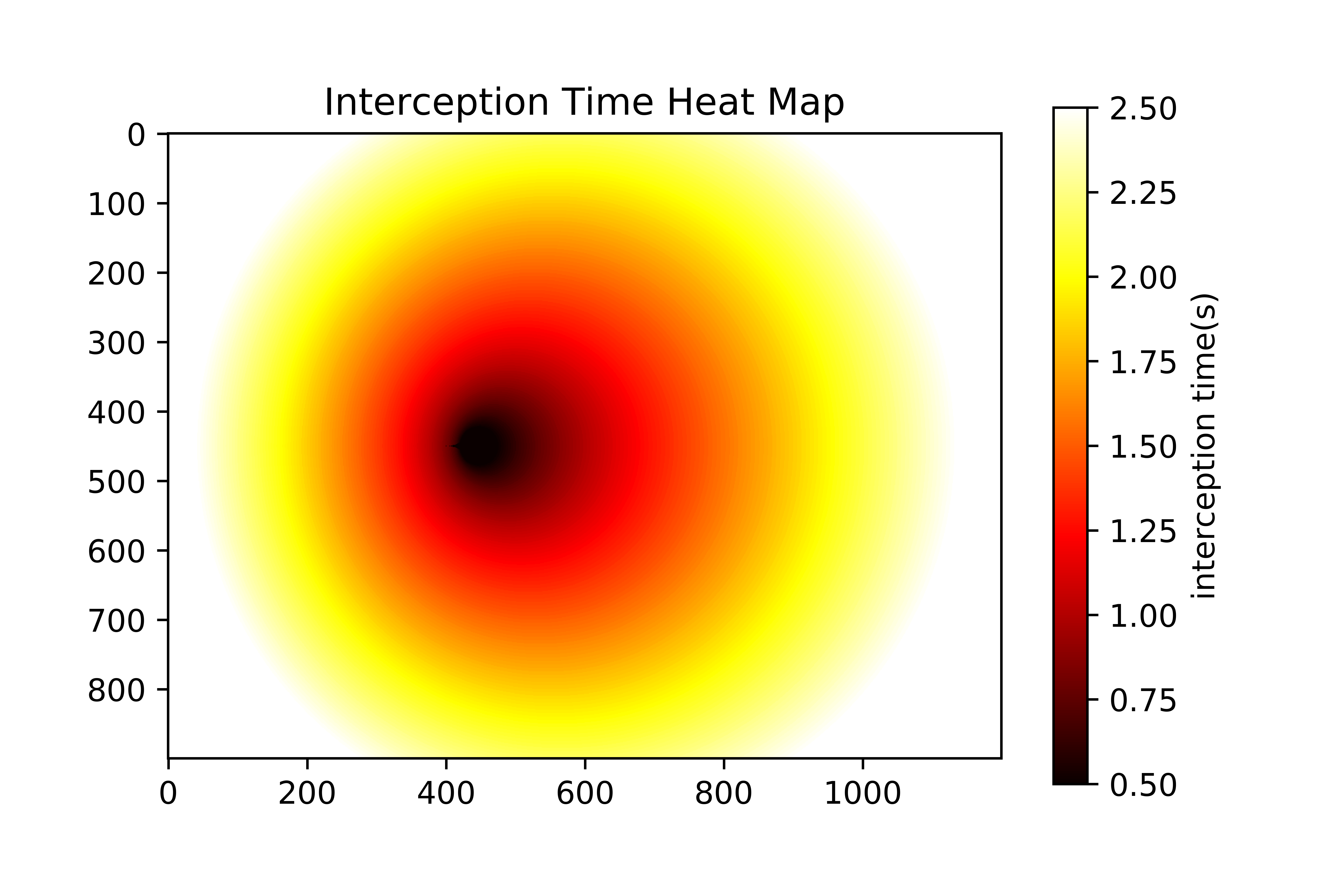}
  \caption{1m/s Ball Speed Interception Time Heat Map}
  \label{inter_1}
\end{minipage}%
\begin{minipage}{.5\textwidth}
  \centering
  \includegraphics[height=1.5in]{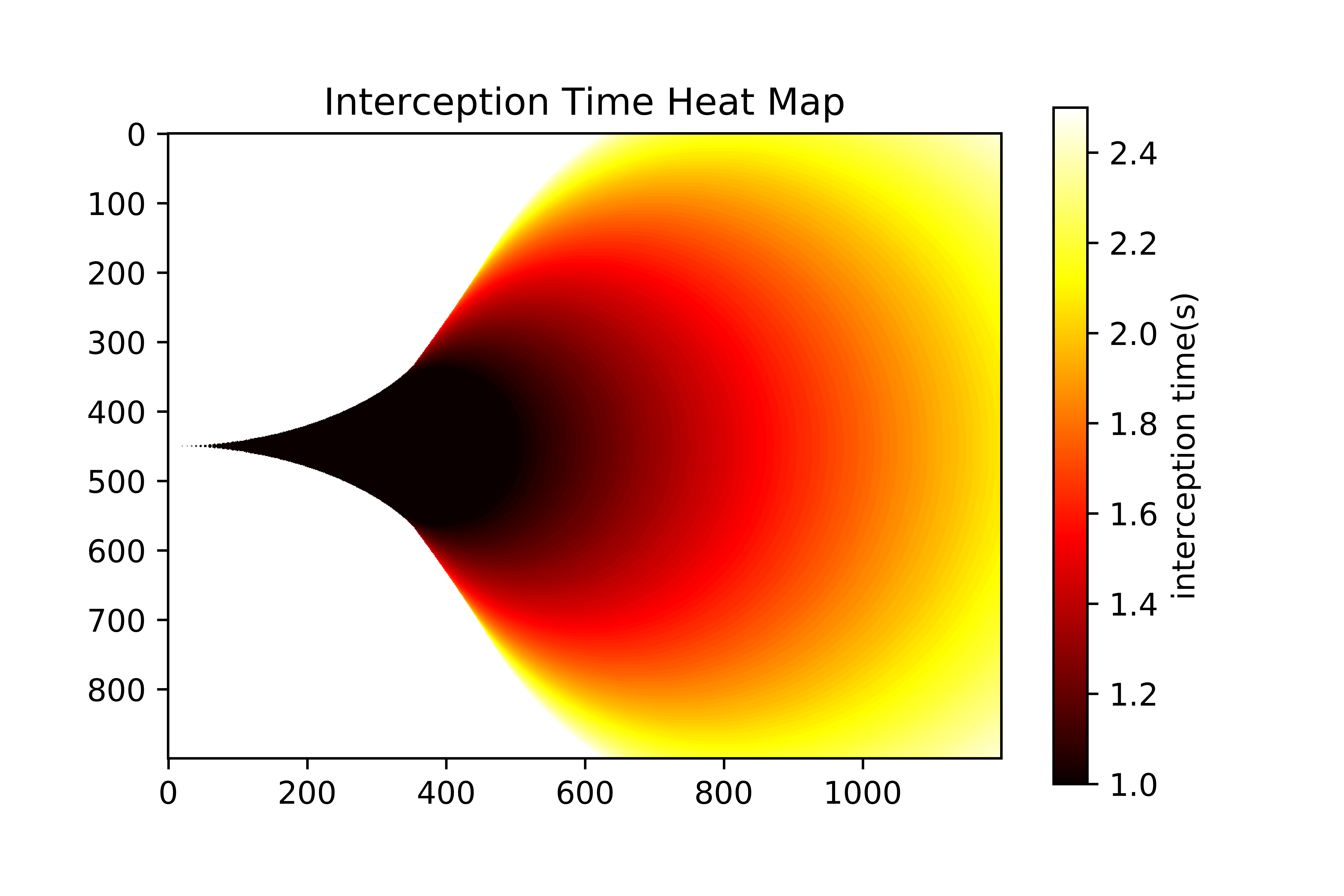}
  \caption{4m/s Ball Speed Interception Time Heat Map}
  \label{inter_2}
\end{minipage}
\end{figure}

Fig. \ref{inter_1} and Fig. \ref{inter_2} shows the interception time of a stationary robot at different positions under two different ball speeds. Darker areas represent shorter interception time, while lighter areas represent longer interception time. In Fig. \ref{inter_1}, the initial position of the ball is $(400 cm, 450 cm)$, and ball speed is low ($1m/s$), so at a certain time, the closer the robot gets to the ball, the less time it has to intercept the ball. However, when ball speed is high, it has different conclusion. In Fig. \ref{inter_2}, the initial position of the ball is $(0 cm, 450 cm)$, and ball speed is high ($4m/s$). Robot cannot intercept the ball in most places on the left side, and there is an obvious boundary. If the position of the robot is within the boundary(i.e. the dark area), it can intercept the ball in a short time, but if not, it will cost much, and may never intercept the ball before it out of the field. In the old saying of China, it is called ``A little error may lead to a large discrepancy''.

\begin{algorithm}
\caption{Search-Based Interception Prediction}
\label{alg}
\begin{algorithmic}
\REQUIRE $\Delta t$, ball initial position $P_0$ and velocity $v_0$, robot initial position $P_r$ and velocity $v_r$
\STATE $k \leftarrow 0$
\REPEAT
\STATE $P_k \leftarrow predictBallPosition(P_0, v_0, k\Delta t)$
\STATE $T_k \leftarrow predictRobotArrivalTime(P_r, v_r, P_k)$
\STATE $k \leftarrow k + 1$
\UNTIL $T_k \leq k\Delta t$ or $P_k$ out of the field
\STATE $P_{best} \leftarrow P_k$
\STATE $T_{best} \leftarrow T_k$
\end{algorithmic}
\end{algorithm}

In \cite{zju_2014} we developed a ``FSM-based Role Matching'' mechanism, using the square of the distance between the current positions of the robots and expected roles’ target positions as the cost function. Considering the above situation, it is actually wrong when math robots to intercept the ball if we choose the ball position as the target position. A better way is using the time that robots move from the current positions to the target as a loss function, and if the target is a ball, using Algorithm. \ref{alg} can match an optimal robot to get the ball, that will improve our ball possession rate.

\subsection{Implementation of Intercept Ball Skill}
\begin{figure}[h!t]
\centering
\begin{minipage}{.5\textwidth}
  \centering
  \includegraphics[height=1.5in]{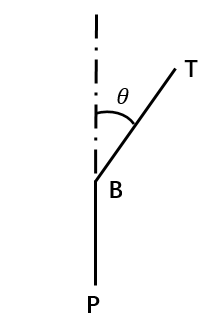}
  \caption{Intercept Ball Skill Sketch Map}
  \label{zget}
\end{minipage}%
\begin{minipage}{.5\textwidth}
  \centering
  \includegraphics[height=1.5in]{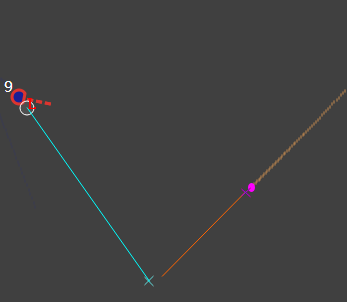}
  \caption{Actual Effect of Intercept Ball Skill}
  \label{zget_result}
\end{minipage}
\end{figure}

Since we can predict robot optimal interception position, we developed a dynamic interception skill based on our robot location, optimal interception location, and kicking location. As shown in the Fig. \ref{zget}, assuming that the location of our robot is location $P$, the predicted optimal interception location is location $B$, the kicking location is location $T$, and the Angle between $PB$ and $BT$ is $\theta$, we choose different interception methods according to the absolute value of $\theta$. If the absolute value of $\theta$ is less than $45$ degrees, we select the ``Chase Ball'' skill, which is chasing the ball forward and kicking to the target location; If it is between $45$ and $120$ degrees, we select the ``Intercept Ball'' skill. We intercept the ball, then turn around and shoot at the target location. If it is greater than $120$ degrees, we select the ``Touch Ball'' skill, run to the ball and kick the ball directly to the target location.

\subsection{Implementation of Marking Skill}
Based on the implementation of the interception prediction algorithm, we have developed a new ``Marking'' skill based on the best interception location and the shortest interception time of both sides.

\begin{figure}[h!t]
  \centering
  \includegraphics[height=2in]{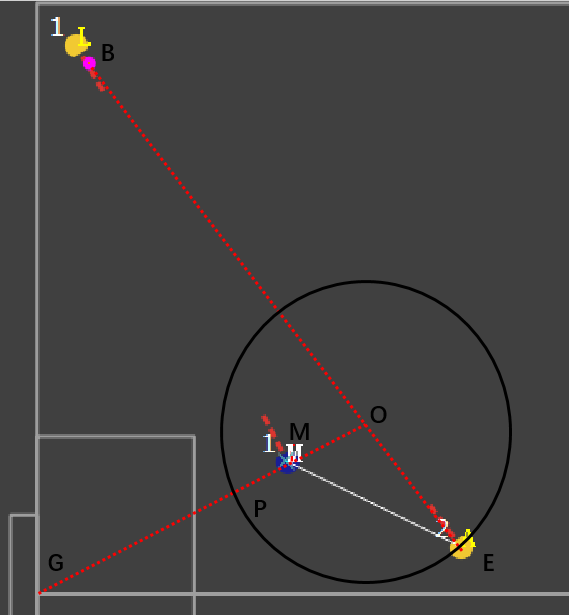}
  \caption{Marking Skill Sketch Map}
  \label{zmarking}
\end{figure}

Fig. \ref{zmarking} shows an application of this skill. Let us assume that we are on the blue side, the yellow robot No.1 is now controlling the ball, location $B$ is the location of the ball, the yellow robot No.2 is the possible ball catching robot, location $E$ is its location, and $G$ is the center location of the goal. Since the robot generally moves at a low speed when passing and catching the ball, we can assume that the robot on the field is stationary. We assume that the ball will move along ray $BE$ at the maximum speed from location $B$ at the next moment, and the yellow robot No.2 will intercept the ball according to the maximum movement ability of the enemy. According to the interception prediction algorithm, the optimal interception location of the yellow robot No.2 can be calculated at location $O$. Draw a circle $O$ with location $O$ as the center of the circle and the length of segment $OE$ as the radius. Assuming that our robot has the same movement ability as the enemy robot, as long as our robot No.1 is in circle $O$, it can intercept the ball before the yellow robot No.2. In order to balance the grab and defense, we line segment $OG$, and segment $OG$ intersects the circle $O$ at location $P$. We will choose the position like $M$ of our No.1 robot station on the segment $OP$ near the location $P$.

The application of this skill allows us more likely to grab the ball passing of the other side, thus greatly improving our ball possession rate.

\end{document}